\def\year{2017}\relax
\documentclass[letterpaper]{article} 
\usepackage{aaai17}  
\usepackage{times}  
\usepackage{helvet}  
\usepackage{courier}  
\usepackage{url}  
\usepackage{graphicx}  
\begin{document}
%
\title{Interactive Learning of State Representation through Natural Language Instruction and Explanation}
\author{Qiaozi Gao, Lanbo She, and Joyce Y. Chai\\
Department of Computer Science and Engineering\\
Michigan State University, East Lansing, MI 48824, USA\\
\{gaoqiaoz, shelanbo, jchai\}@cse.msu.edu\\
}
\maketitle

\begin{abstract}
One significant simplification in most previous work on robot learning is the closed-world assumption where the robot is assumed to know ahead of time a complete set of predicates describing the state of the physical world. However, robots are not likely to have a complete model of the world especially when learning a new task. To address this problem, this extended abstract gives a brief introduction to our on-going work that aims to enable the robot to acquire new state representations through language communication with humans.
\end{abstract}

\section{Introduction}
As cognitive robots start to enter our lives, being able to teach robots new tasks through natural interaction becomes important~\cite{matuszek2012joint,liu16a,liu16b,chai2017}. One of the most natural ways for humans to teach task knowledge is through natural language instructions, which are often expressed by verbs or verb phrases. Previous work has investigated how to connect action verbs to low-level primitive actions \cite{branavan2009reinforcement,mohan2014learning,she14sigdial,misra2015environment,misra2016tell,she2016incremental,she2017interactive}. In most of these studies, a robot first acquires the state change of an action from human demonstrations and represents verb semantics using the desired goal state. With learned verb semantics, given a language instruction, the robot can apply the goal states of the involved verbs to plan for a sequence of low-level actions.

\begin{figure}
\includegraphics[width=0.47\textwidth]{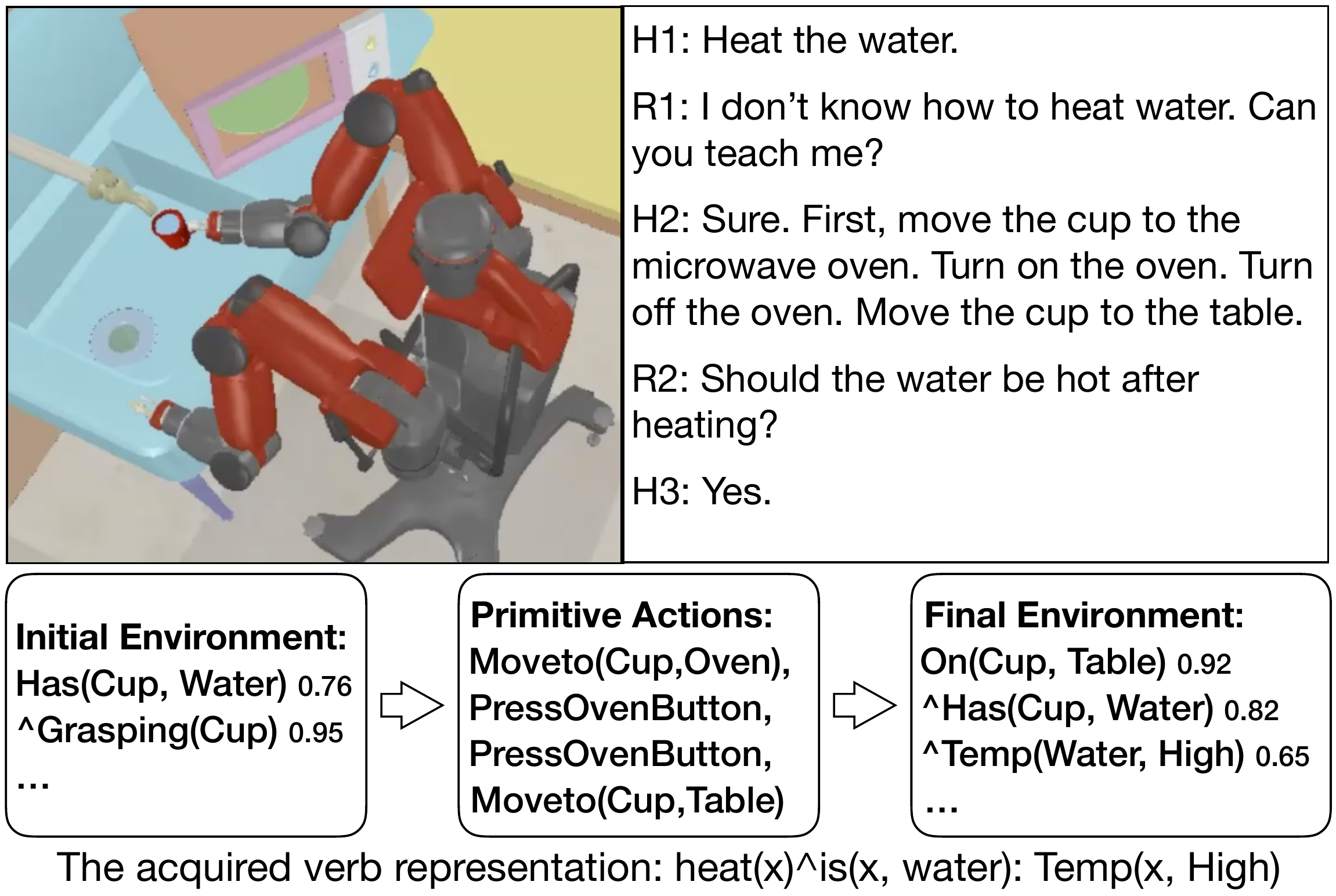}
\centering
\vspace{-10pt}
\caption{An example of learning the state-based representation for the command {\em``heat water''}.}
\label{fig:boil}
\vspace{-5pt}
\end{figure}

For example, a human can teach the robot the meaning of the verb phrase {\em ``heat water''} through step-by-step instructions as shown in H2 in Figure~\ref{fig:boil}. The robot can identify the state change by comparing the final environment to the initial environment. The learned verb semantics is represented by the goal state (e.g., \texttt{Temp(x,High)}). To handle uncertainties of perception, the robot can also ask questions and acquire better representations of the world through interaction with humans~\cite{she2017interactive}.

Previous work is developed based on a significant simplification: the robot knows ahead of time a complete set of predicates (or classifiers) that can describe the state of the physical world. However in reality robots are not likely to have a complete model of the world.
Thus, it is important for the robot to be proactive~\cite{chai2014,chai2016} and transparent~\cite{alexandrova2014,alexandrova2015,whitney2016,hayes2017} about its internal representations so that humans can provide the right kind of feedback to help capture new world states. To address this problem, we are developing a framework that allows the robot to acquire new states through language communication with humans.  

\section{Interactive State Acquisition}

\begin{figure}
\includegraphics[width=0.47\textwidth]{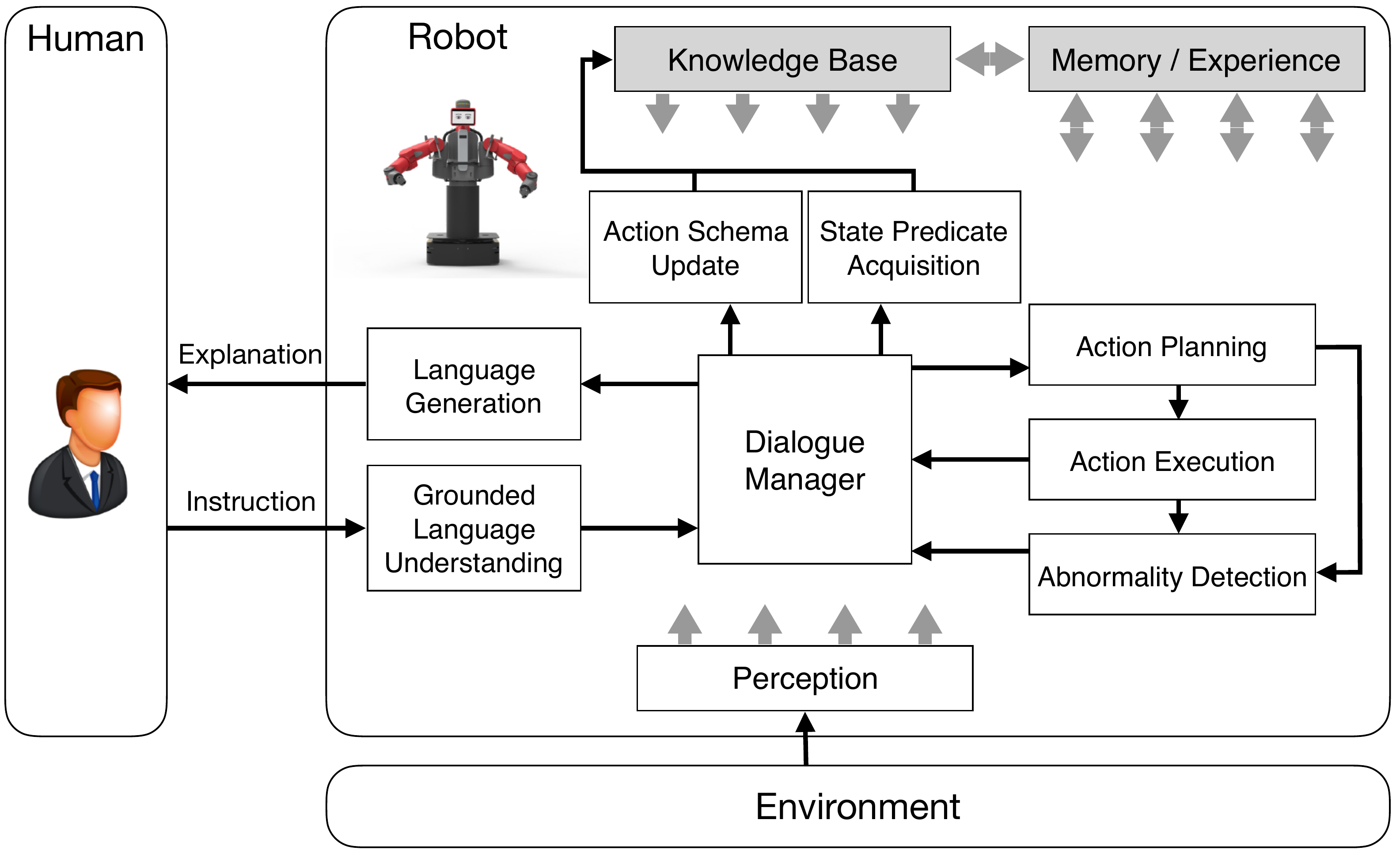}
\centering
\vspace{-10pt}
\caption{Interactive acquisition of new physical states.}
\label{fig:process}
\vspace{-5pt}
\end{figure}

The proposed framework is shown in Figure~\ref{fig:process}. In additional to modules to support language communication (e.g., {\bf grounded language understanding} and {\bf dialogue manager}) and action (e.g., {\bf action planning} and {\bf action execution}), the robot has a {\bf knowledge base} and a {\bf memory/experience} component. The {\bf knowledge base} contains the robot's existing knowledge about verb semantics, state predicates, and action schema (both primitive actions and high-level actions). The {\bf memory/experience} component keeps track of interaction history such as language input from the human and sensory input from the environment. 

\begin{figure}
\includegraphics[width=0.48\textwidth]{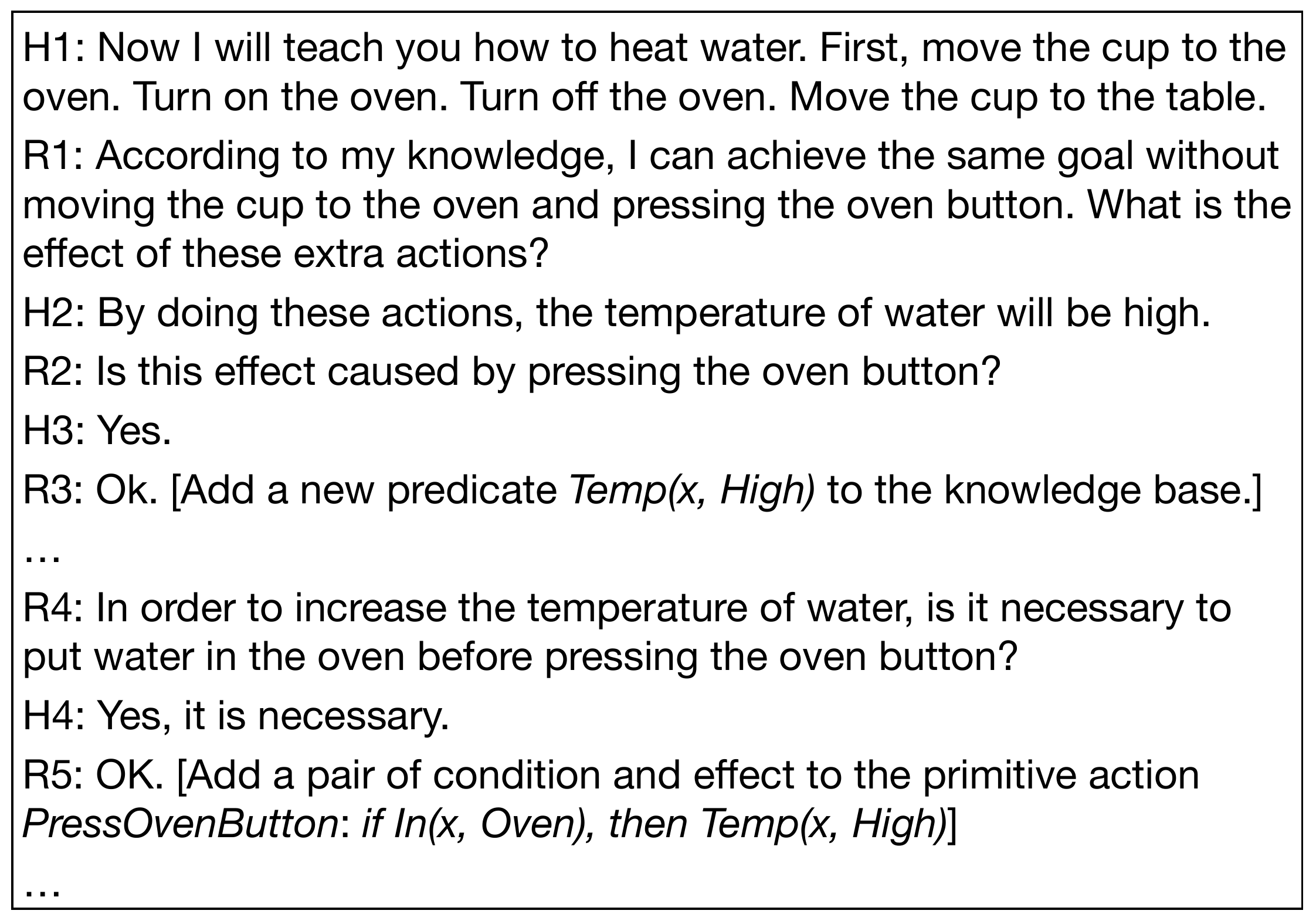}
\centering
\vspace{-10pt}
\caption{An example of interactively learning a new state predicate during the human teaches the robot how to  ``heat water''.}
\label{fig:dialogue}
\vspace{-5pt}
\end{figure}

Suppose the robot does not have the state predicate \texttt{Temp(x, High)} in its knowledge base and the effect of the primitive action \texttt{PressOvenButton} only describes the change of the oven status (i.e., \texttt{Status(Oven, On)}). Our framework will allow the robot to acquire the new state predicate \texttt{Temp(x, High)} and update action representation (shown below with the added condition and state in bold) through interaction with the human as shown in Figure~\ref{fig:dialogue}.  
\\
\begin{scriptsize}
\indent \indent if (not Status(Oven, On)), then:\\
\indent \indent \indent Status(Oven, On) {\bf and if In(x, Oven), then: {\bf Temp(x, High)}} \\ 
\indent \indent if Status(Oven, On), then:\\
\indent \indent \indent not Status(Oven, On)\\
\end{scriptsize}
This framework includes two main processes: (1) acquiring and detecting new states; and (2) updating action representation.




\subsection{Acquiring and Detecting New States}




Since an incomplete action schema can cause planning problems~\cite{gil1994learning}, the robot can potentially discover the related abnormality by retrospective planning. In our example, the robot does not have the state predicate \texttt{Temp(x,High)} in its current knowledge base. Thus in the robot's mind, the final environment will not contain \texttt{Temp(Water, High)}. After the human provides instructions on how to heat water, the dialogue manager calls a retrospective planning process based on the robot's current knowledge to achieve the final environment. Then the {\bf abnormality detection} module compares the planned action sequence with human provided action sequence and finds that the planning result lacks of primitive actions \texttt{Moveto(Cup, Oven)} and  \texttt{PressOvenButton}. Once an abnormality is detected, the robot explains its limitation to human for diagnosis (R1). Note that there is a gap between the robot's mind and the human's mind. The human does not know the state predicates that the robot uses to represent the physical world. In order for humans to understand its limitation, the robot explains the differences between the two action sequences, and requests the human to provide missing effects. Based on the human's response, the {\bf state predicate acquisition} module adds a new state predicate \texttt{Temp(x, High)} to the knowledge base.
Next the robot needs to know how to detect such state from the physical environment. State detection is a challenging problem by itself. It often involves classifying continuous signals from the sensors into certain classes, for examples, as in previous work that jointly learns concepts and their physical groundings by integrating language and vision~\cite{matuszek2012joint,krishnamurthy2013jointly}. We are currently exploring approaches that automatically bootstrap training examples from the web for detection of state. 


\subsection{Updating Action Representation}


Once a new state predicate is acquired, the robot needs to know what primitive actions and under what conditions the related state change can be caused. 
The relevant primitive action can be identified by applying the state detection model to the sensory input from the environment that is stored in the memory. Now the problem is reduced to determine what condition is needed to cause that particular state change. And this is similar to the {\it planning operator acquisition} problem, which has been studied extensively~\cite{wang1995learning,amir2008learning,Mourao12,zhuo2014action}. However, in previous work, primitive actions are acquired based on multiple demonstration instances. Inspired by recent work that support interactive question answering~\cite{cakmak2012,she2017interactive}, we intend to enable robots to ask questions to identify the correct conditions for primitive actions (R4). We are currently extending an approach based on reinforcement learning to learn when to ask what questions. Based on the human's response, the {\bf action schema update} module adds a pair of condition and effect to the primitive action \texttt{PressOvenButton} as shown earlier. 

\section{Conclusion and Future Work}

This paper gives a brief introduction to our on-going work that enables the robot to acquire new state predicates to better represent the physical world through language communication with humans. Our current and future work is to evaluate this framework in both offline data and real-time interactions, and extend it to interactive task learning.

\section{Acknowledgment}
This work was supported by the National Science Foundation (IIS-1208390 and IIS-1617682) and the DARPA XAI program under a subcontract from UCLA (N66001-17-2-4029).

\bibliographystyle{aaai}
\bibliography{inuse}

\end{document}